%% file: New_IEEEtran_how-to.tex
\documentclass[lettersize,journal]{IEEEtran}
\usepackage{amsmath,amsfonts}
\usepackage{algorithm}
\usepackage{algpseudocode}
\usepackage{array}
\usepackage{booktabs}  
\usepackage{multirow} 
\usepackage[caption=false,font=normalsize,labelfont=sf,textfont=sf]{subfig}
\usepackage{textcomp}
\usepackage{stfloats}
\usepackage{url}
\usepackage{verbatim}
\usepackage{xcolor}
\usepackage{graphicx}
\def\BibTeX{{\rm B\kern-.05em{\sc i\kern-.025em b}\kern-.08em
    T\kern-.1667em\lower.7ex\hbox{E}\kern-.125emX}}
\usepackage{balance}
\begin{document}
\title{MaskHOI: Robust 3D Hand-Object Interaction Estimation via Masked Pre-training}
\author{Yuechen Xie,
        Haobo Jiang,
        Jian Yang,
        Yigong Zhang,
        and Jin Xie

\thanks{Manuscript received Month Day, Year; revised Month Day, Year. This work was supported in part by... (). (Corresponding author: Jin Xie.)}
\thanks{Yuechen Xie and Jin Xie are with the PCA Lab, School of Intelligence Science and Technology, Nanjing University, Nanjing 210093, China (e-mail: yuechen.xie@qq.com; csjxie@nju.edu.cn).}%
\thanks{Haobo Jiang is with Nanyang Technological University, Singapore 639798 (e-mail: haobo.jiang@ntu.edu.sg).}%
\thanks{Jian Yang is with the PCA Lab, Nanjing University of Science and Technology, Nanjing 210094, China (e-mail: csjyang@njust.edu.cn).}%
\thanks{Yigong Zhang is with the PCA Lab, Nankai University, Tianjin 300071, China (e-mail: 9820220110@nankai.edu.cn).}%
}
\markboth{Journal of \LaTeX\ Class Files,~Vol.~18, No.~9, September~2020}%
{How to Use the IEEEtran \LaTeX \ Templates}

\maketitle

\input{section/0_abstract.tex}

\input{section/1_intro.tex}

\input{section/2_related.tex}

\input{section/3_method.tex}

\input{section/4_exp.tex}

\input{section/6_conclusion.tex}

\bibliographystyle{IEEEtran}  
\bibliography{references}

\end{document}

%% file: section/0_abstract.tex
\begin{abstract}
    In 3D hand-object interaction (HOI) tasks, 
    estimating precise joint poses of hands and objects from monocular RGB input remains highly challenging 
    due to the inherent geometric ambiguity of RGB images and the severe mutual occlusions that occur during interaction.
    To address these challenges, we propose MaskHOI, 
    a novel Masked Autoencoder (MAE)–driven pretraining framework for enhanced HOI pose estimation. 
    Our core idea is to leverage the masking-then-reconstruction strategy of MAE to encourage the feature encoder to infer missing spatial and structural information, 
    thereby facilitating geometric-aware and occlusion-robust representation learning. 
    Specifically, based on our observation that human hands exhibit far greater geometric complexity than rigid objects, 
    conventional uniform masking fails to effectively guide the reconstruction of fine-grained hand structures. 
    To overcome this limitation, 
    we introduce a Region-specific Mask Ratio Allocation, 
    primarily comprising the region-specific masking assignment and the skeleton-driven hand masking guidance. 
    The former adaptively assigns lower masking ratios to hand regions than to rigid objects, 
    balancing their feature learning difficulty,
     while the latter prioritizes masking critical hand parts (e.g., fingertips or entire fingers) to realistically simulate occlusion patterns in real-world interactions. 
    Furthermore, to enhance the geometric awareness of the pretrained encoder, we introduce a novel Masked Signed Distance Field (SDF)-driven multimodal learning mechanism. 
    Through the self-masking 3D SDF prediction, 
    the learned encoder is able to perceive the global geometric structure of hands and objects beyond the 2D image plane,
    overcoming the inherent limitations of monocular input and alleviating self-occlusion issues.
    Extensive experiments demonstrate that our method significantly outperforms existing state-of-the-art approaches.
\end{abstract}

%% file: section/1_intro.tex
\section{Introduction}
\label{sec:intro}

Achieving accurate and robust estimation of 3D hand-object interaction (HOI)  scenes from monocular RGB images 
is a significant and challenging research topic in the field of computer vision.
However,  HOI estimation from monocular RGB input faces two core challenges.
First, monocular vision lacks direct 3D geometric depth information, 
forcing the system to rely primarily on 2D texture features for indirect inference, 
which introduces uncertainty in understanding the scene's 3D structure.
Second, the prevalent occlusion problem in HOI scenarios is particularly pronounced.
This occlusion includes not only mutual occlusion between the hand and object but also complex self-occlusions of the hand and potential self-occlusions of the object due to its structure.
These multi-layered, 
dynamically changing occlusions can lead to significant loss of critical visual information, 
making it extremely challenging to accurately infer the complete 3D interaction state from the remaining, 
often partial and possibly unclear 2D observations, 
thus posing a severe test for the robustness of algorithms.
Therefore, 
robustly understanding and estimating the 3D poses and interaction relationships between hands and objects under 
incomplete information is a core issue that urgently needs to be addressed in this field.
\begin{figure}[!tbp]
    \centering
    \includegraphics[width=1.0\columnwidth]{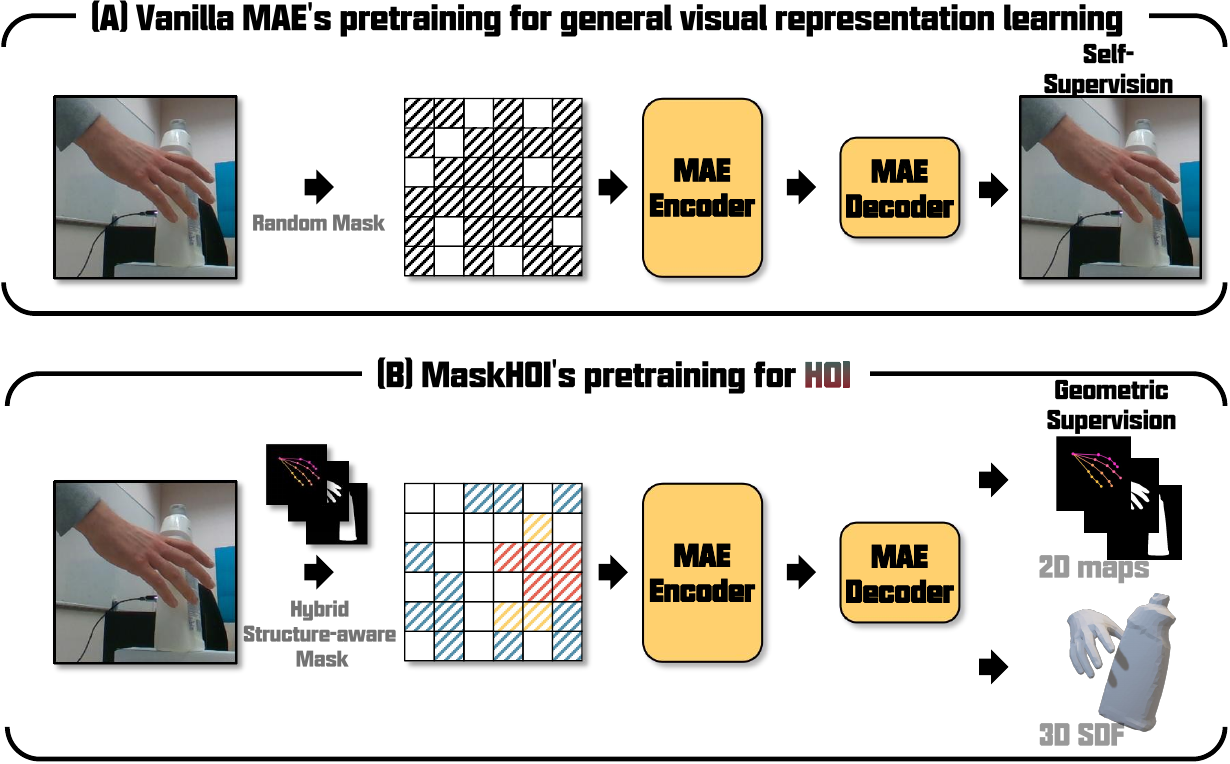}
    \caption{Comparison of vanilla MAE\cite{he2022masked} with our MaskHOI.
    Different from the vanilla MAE, 
    which focuses on reconstructing generic image content, 
    our MaskHOI is designed to enhance the encoder's robustness by focusing on reconstructing target-specific 3D geometric information.
    This allows the encoder to effectively capture robust feature representations even in HOI scenarios with complex occlusion variations.  
    }
    \label{fig:intro}
\end{figure}

To address these challenges, existing research has made some valuable explorations. 
For instance, 
some methods recognize the importance of leveraging 3D geometric priors and attempt to introduce representations like 3D Signed Distance Fields (SDF) 
as a bridge between 2D observations and 3D geometric space\cite{qi2024hoisdf,chen2022alignsdf}.
The use of SDF allows the model to better understand and infer the geometric shapes of partially occluded objects or hands, 
alleviating the issue of incomplete geometric information caused by factors like self-occlusion under monocular input to some extent.
At the same time, 
some studies have observed that even the occluded areas (usually referring to the surface of the object) are strongly correlated with the occluded hand\cite{park2022handoccnet}.
However, despite the progress made by introducing 3D priors like SDF and the information reuse strategies for occluded areas in methods like HandOccNet\cite{park2022handoccnet}, 
there is still a lack of comprehensive and effective solutions in the field to address the two core challenges of mutual occlusion and self-occlusion that are prevalent in complex interaction scenarios.
As an example, 
the HandOccNet method fundamentally relies on the quantity and quality of visible hand information. 
However, 
in occluded environments, 
effective hand information is often scarce, 
and the quality of available reference information can significantly decrease.

Therefore, 
based on the above analysis,
we propose MaskHOI, a novel Masked Autoencoder (MAE)–driven pretraining framework for enhanced HOI pose estimation,
The core idea of MaskHOI is to leverage the masking-then-reconstruction strategy of MAE to encourage the feature encoder to infer missing spatial and structural information, 
thereby facilitating geometric-aware and occlusion-robust representation learning.

Specifically, we employ the self-masking learning mechanism of MAE in the MIM paradigm\cite{he2022masked}, 
explicitly masking a significant portion of the effective information during the pretraining process.
This mechanism force the encoder to infer complete HOI scene information from incomplete and missing image information during the self-masking pretraining process.
By doing so, the encoder is compelled to learn to restore and predict specific 3D information related to HOIs from incomplete image information.

Notably,
unlike traditional MAE-related methods trying to reconstruct generic image content,
MaskHOI utilizes the self-masking mechanism as a challenge of information loss,
guiding the encoder to focus on restoring and predicting specific 3D information related to the HOI itself,
rather than reconstructing generic image content that does not distinguish between foreground and background, as shown in Figure~\ref{fig:intro}.
At the same time,
we observe that the geometric complexity of hands is much higher than that of rigid objects,
and the uniform random masking strategy in traditional MAE often fails to effectively guide the reconstruction of fine-grained hand details.
Therefore, in contrast to the random masking strategy used in traditional MAE for generic visual representation learning,
we design a Hybrid Structure-aware Masking Strategy tailored for the HOI task and the fundamental differences in motion structure and shape texture variations between hands and objects.
Specifically, to address the differences between hands and objects,
we introduce Region-specific Masking Rate Allocation.
This masking strategy allocates region-specific masking rates based on the difficulty of target reconstruction,
ensuring that the model applies different learning pressures to different targets when inferring the overall scene information of HOIs.
Furthermore, we introduce a Skeleton-guided masking sampling strategy to specifically optimize feature learning for hands in HOI scenarios.
This strategy prioritizes masking certain key areas of the hand (e.g., fingertips, complete fingers) through structured masking sampling to simulate occlusion patterns in real interaction scenarios,
encouraging the encoder to effectively capture robust features even in HOI scenes with frequent occlusions.
At the same time, we note that while our method can drive the encoder to obtain robust feature representations in occluded scenes,
the self-masking mechanism alone cannot address the perception limitations of 2D encoders under monocular input.
This is particularly true for non-rigid targets like hands, 
where self-occlusion patterns in HOI scenarios are often highly complex and dynamic.
Therefore, how to construct an explicit representation of global 3D information is an urgent problem to be solved.
To this end,
we propose a multi-modal learning approach based on Masked Signed Distance Field (SDF),
which aims to construct a global geometric-aware pretraining to enhance the encoder's perception capabilities for occlusion and self-occlusion under monocular views.
Specifically, in the MaskHOI's MAE-based masked pretraining process,
we introduce the SDF modality as an additional modality,
and learn to reconstruct the SDF of hands and objects in HOI scenes from incomplete tokens.
The advantage of this modality is that it can provide global 3D information beyond the limitations of monocular views and 2D image representations.
The combination of reconstructing this global geometric information and the masking mechanism 
can effectively alleviate self-occlusion and occlusion issues under monocular input.
Furthermore, considering that the SDF modality typically describes only the external geometric shape of HOIs and lacks descriptions of the object's pose and the hand's internal structure,
we directly introduce a parametric supervision signal at the feature level during the MaskHOI masked pretraining process, 
rather than regressing parameters through indirect geometric representations\cite{Wang_2021_CVPR,hu2020single}.
This direct connection between tokens and the supervision signal encourages the encoder to learn the inherent human and kinematic priors of the MANO model\cite{romero2017embodied},
avoiding the generation of visually similar but geometrically unreasonable hand shape or object poses.
At the same time, 
in downstream tasks, 
the prior knowledge learned by the encoder can also guide subsequent decoding networks to perform more accurate pose estimation or reconstruction.

In summary, our contributions can be summarized as follows:
\begin{itemize}
    \item We propose MaskHOI, a novel HOI task-specific masked pretraining paradigm, 
    which leverages the masking-then-reconstruction mechanism improve the encoder's perceiving capabilities under the condition of frequent occlusion in HOI scenes.
    \item We design a Hybrid Structure-aware Masking Strategy that introduces Region-specific Masking Rate Allocation and Skeleton-guided masking sampling to balance the learning difficulty of hands and objects in HOI tasks.
    \item We introduce a multimodal pretraining approach that incorporates Masked Signed Distance Field (SDF) learning to enhance the encoder's perception capabilities for both occlusion and self-occlusion in HOI scenarios.
    \end{itemize}

Extensive experiments demonstrate that our method significantly outperforms existing state-of-the-art methods in HOI pose estimation tasks.

%% file: section/2_related.tex
\section{Related work}
\label{sec:related}
\subsection{Hand-object interaction}

Hand-object interaction (HOI) research plays a central role in advancing various technological fields, including virtual reality (VR), augmented reality (AR)~\cite{chen2019overview}, gesture recognition, human-computer interaction (HCI)\cite{ren2020review}, and robotics\cite{billard2019trends}. 
Accurately reconstructing the 3D poses of hands and objects, along with their complex interactions, is crucial for understanding and mimicking intricate human behaviors. 
This leads to more natural and intuitive control in immersive digital environments and facilitates advanced robotic manipulation tasks.

However, 3D hand-object interaction research faces multiple fundamental challenges.
Among these, severe occlusion is a common and recurring issue in 3D hand and object mesh estimation, especially in monocular images, where hands and objects often severely occlude each other.
This problem is at the core of nearly all related studies and directly affects feature extraction and pose accuracy.
Secondly, the high joint flexibility and inherent pliability of human hands introduce significant complexity, making pose configurations extremely challenging, with considerable variations in hand shape and size.
Therefore, accurate capture of the intricate details of the interactions ensures physical plausibility.
\subsubsection{Solving Occlusion in Hand-object Pose Estimation}
The occlusion problem in the estimation of hand-object poses has been a long-standing challenge in the field.
Early 3D hand mesh estimation methods often tended to ignore information in occluded regions, treating them as noise.
However, recent studies have strongly pointed out that these occluded regions have a strong association with the hand and object, providing very useful information for complete 3D hand mesh estimation.
For example, HandOccNet\cite{park2022handoccnet} aims to fully utilize information within occluded regions to enrich image features for 3D hand mesh estimation.
Its subsequent work, HandGCAT\cite{HandGCAT2023}, further emphasizes the use of prior knowledge of the hand as compensatory information to improve characteristics in occluded regions.
It employs Knowledge-Guided Graph Convolution (KGC) modules and Cross-Attention Transformers (CAT)\cite{vaswani2023attentionneed} modules to achieve this goal.
HFL-Net addresses the issue of "competitive feature learning" that arises in occluded scenarios, where a single-backbone network struggles to extract features for both the hand and object effectively.
Chen et al.\cite{9390307} introduce a pyramid architecture of multibranch features that uniquely leverages the inherent geometric stability of objects as prior knowledge.
By using object features as key/value inputs to the Transformer, it guides the reconstruction of occluded hand regions, effectively enhancing hand features and improving model robustness in complex occlusion scenarios.
Keypoint-Transformer\cite{hampali2022keypoint} addresses the challenges of reconstructing the 3D hand-object interaction pose by explicitly separating joint localization from joint identification/association tasks.
Jiang et al. \cite{jiang2023se} transform traditional single-step pose parameter generation into multi-step denoising generation and outlier removal to handle these difficult scenarios.
In general, the evolution of occlusion handling has shifted from passive treatment of occluded regions (treating them as noise) to actively leveraging contextual information within these regions.

\subsubsection{Geometric Representation of Hands and Objects}
Previous methods\cite{Zhang2022DifferentiableSpatialRegression,liang2014parsing} achieved the estimation of hand pose by establishing a direct mapping between the hand key points and the features of the 2D image.

Parametric meshes are a common approach that utilizes existing parameterized hand models, such as MANO~\cite{romero2017embodied}, and assumes that 3D object models are known or available.
These models benefit from strong prior knowledge about the general structure and kinematics of hands and objects.
However, explicit representations also have inherent limitations, including constrained shape deformations and fixed mesh resolutions.
This can lead to imprecise reconstruction of fine details, especially in critical areas like contact regions between hands and objects, as well as when dealing with generic or unknown objects.
In recent years, implicit representation methods have gradually emerged, leveraging neural networks to learn the geometric shapes of hands and objects.
These methods provide a continuous and differentiable way to represent 3D geometry, encoding shape information directly into network parameters.
For example, methods like Grasping Field\cite{karunratanakul2020grasping}, AlignSDF\cite{chen2022alignsdf}, gSDF\cite{chen2023gsdf}, and HOISDF\cite{qi2024hoisdf} achieve 3D pose estimation by learning implicit surface representations of hands and objects.
These methods can handle more complex shape deformations and demonstrate better flexibility and adaptability when dealing with unknown or generic objects.

\subsection{Masked Image Modeling}
Masked Image Modeling (MIM) is a self-supervised learning approach that trains models to predict missing regions from the unmasked parts of an image.
This method has achieved significant success in the field of computer vision, particularly in tasks such as image classification, object detection, and image segmentation.
Prominent methods such as Masked Autoencoders (MAE)~\cite{he2022masked} and BEiT~\cite{bao2021beit} have demonstrated that by masking a significant portion of an input image and assigning a model to reconstruct missing content, the model learns powerful and robust visual representations.
The core idea is that, to successfully predict the masked patches, the model must develop a holistic understanding of the scene, capturing relationships between visible parts.
This paradigm is particularly relevant to the problem of hand-object interaction, where occlusion is a primary challenge.
The random masking in MIM serves as a form of aggressive data augmentation that simulates severe occlusion scenarios.
By training a model to reconstruct masked regions, it implicitly learns to infer missing information from visible context.
This learned capability to \textit{inpaint} or \textit{hallucinat} visual information can be directly transferred to the task of reconstructing occluded hand and object parts from a single image.
Therefore, leveraging MIM as a pre-training strategy or integrating a MIM-like objective into the training process can significantly enhance a model's robustness to occlusion, leading to more accurate and complete 3D hand-object pose estimations.

%% file: section/3_method.tex
\section{Methods}
\label{sec:method}

We propose MaskHOI, a novel MAE–driven pretraining framework for enhanced HOI pose estimation.
The core idea of MaskHOI is to leverage the masking-then-reconstruction strategy of MAE to encourage the feature encoder to infer missing spatial and structural information, 
thereby facilitating geometric-aware and occlusion-robust representation learning.
In the followings, we describe the details of our MaskHOI framework.
\begin{figure*}[htbp]
    \centering
    \includegraphics[width=0.99\textwidth]{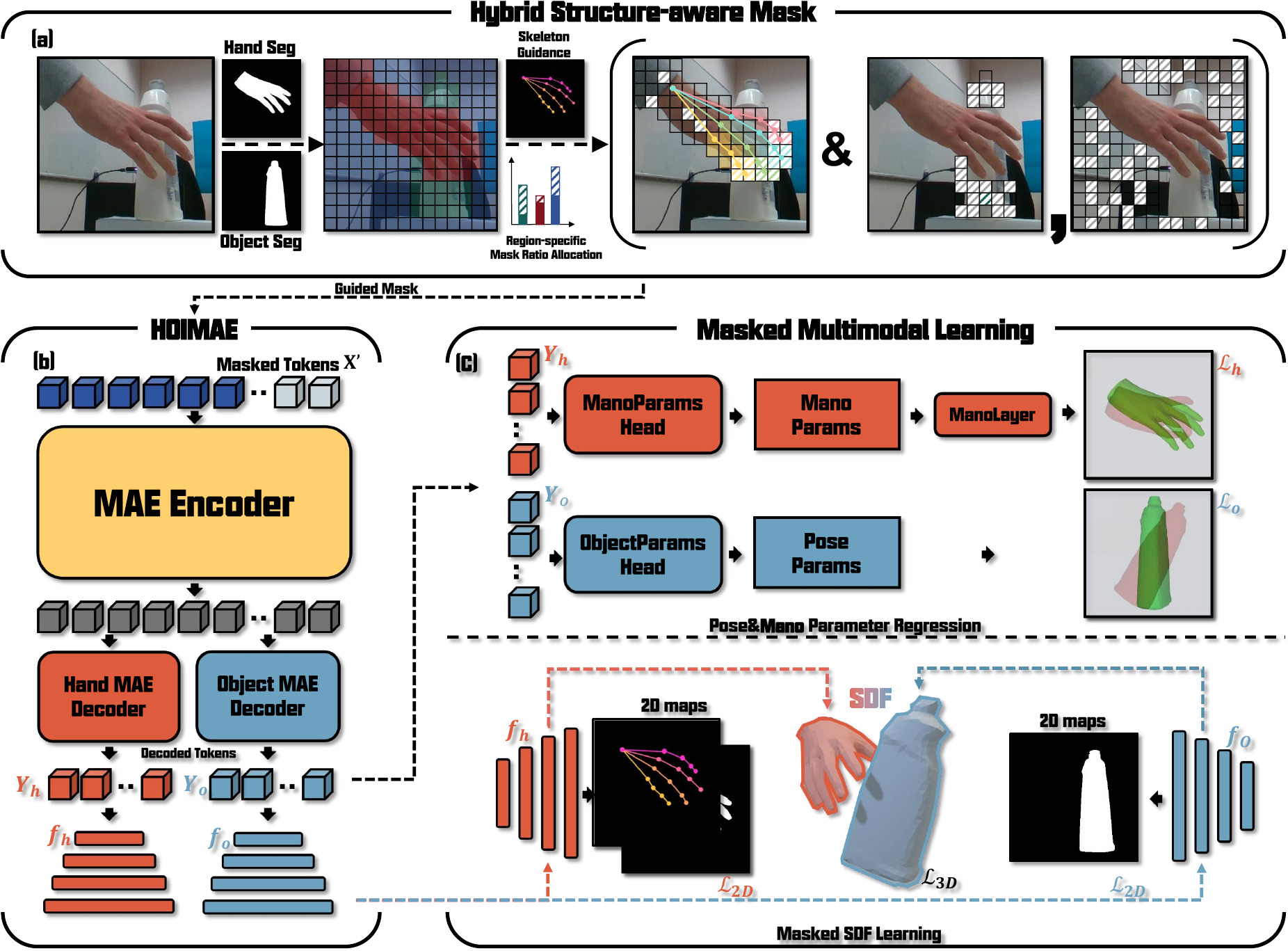}
    \caption{Overview of our MaskHOI framework. MaskHOI composed of three main steps:
    (a)First, we patchify the input image into image tokens. 
     Then, guided from the segmentation masks of hand and object and hand's keypoint localization,
     we apply a Hybrid Structure-aware Masking Strategy to the image tokens to create a subset of unmasked tokens $\textbf{X}'$.
    (b)Second, we feed the unmasked tokens $\textbf{X}'$ into a HOIMAE,
        following the Masked Autoencoder (MAE) paradigm,
        we obtain the recovered $\{\textbf{Y}_h,\textbf{Y}_o\}$ by using a shared encoder and two parallel decoders,
        and obtain multi-scale feature maps $\{\textbf{f}_h,\textbf{f}_o\}$ from the decoded tokens.
    (c)Third, we use the recovered tokens to directly regress the parameters related to the HOI scene.
    Then, segmentation and keypoint heatmaps are also predicted to conducted the multi-modal learning at 2D level.
    Meamwhile, SDF of hand and object are also predicted to conducted the multi-modal learning at 3D level.
    }
    \label{fig:pipeline}
\end{figure*}

\subsection{Overview of MaskHOI}
\label{sec:overview}
MaskHOI framework is designed to enhance the feature learning of the encoder in Hand-Object Interaction (HOI) scenario,
which consists of three stages. 
First, given an input image $\textbf{I}\in \mathbb{R}^{3 \times W \times  H}$,
we first patchify the image into embedding tokens $\textbf{X} \in \mathbb{R}^{N \times D}$, where $N$ is the number of patches and $D$ is the dimension of each patch.
Then, a Hybrid Structure-aware Masking Strategy is applied to the image tokens to create a subset of unmasked tokens $\textbf{X}' \in \mathbb{R}^{N' \times D}$.
Second, we feed the unmasked tokens into a shared transformer-based encoder, and then reconstruct the recovered entire tokens $\{\textbf{Y}_h,\textbf{Y}_o\}$ by two parallel decoders,
multi-scale feature maps $\{\textbf{f}_h,\textbf{f}_o\}$ are obtained at the same time.
Third, we introduce a Geometric-aware Multi-modal Learning approach to conduct a global geometric-aware pretraining by introducing SDF as a additional modality,
in order to enhance the encoder's perceiving ability in HOI scenarios. 
The overall pipeline of our MaskHOI framework is shown in Figure~\ref{fig:pipeline}.

\subsection{Hybrid Structure-aware Masking Strategy}

\begin{figure}[htbp]
    \centering
    \includegraphics[width=0.9\columnwidth]{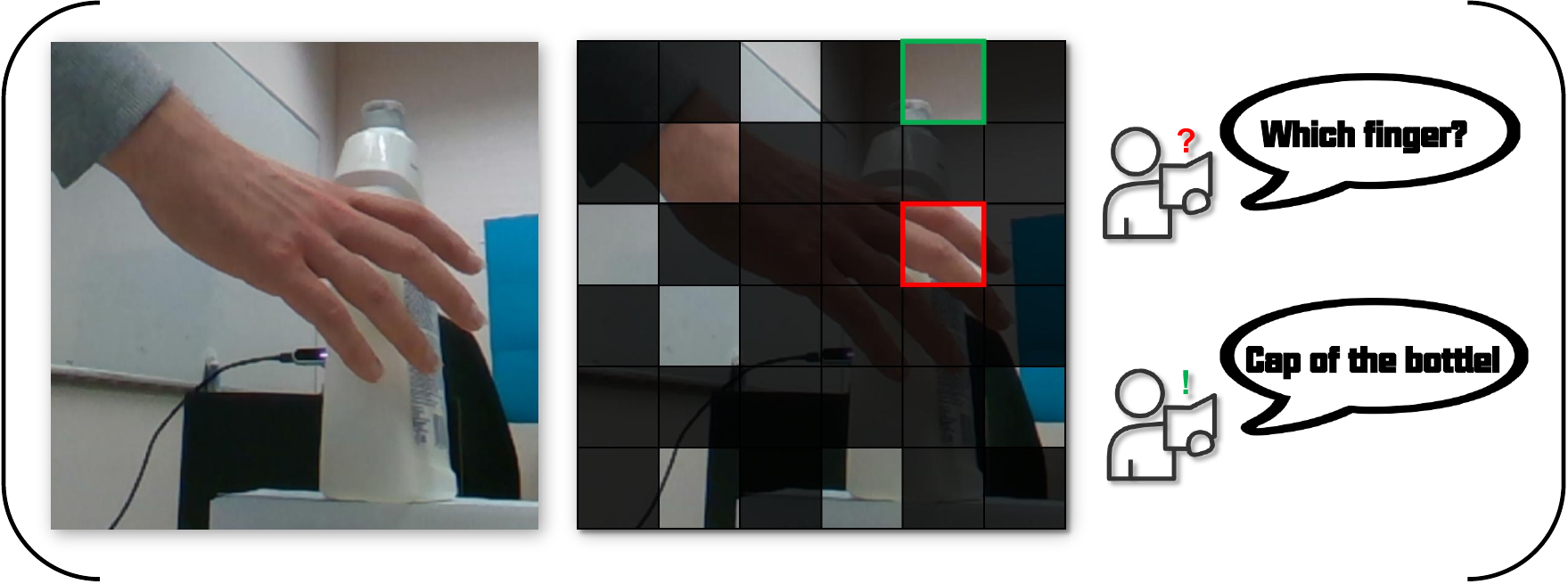}
    \caption{Object and hand under the traditional global random masking strategy.}
    \label{fig:why_mask_stragegy}
\end{figure}
In this section, we explore how to design a reasonable patch masking strategy for both the object and the hand to enable effective masking-then-reconstruction learning in our MaskHOI framework. Our first key insight is that directly applying the random masking strategy (commonly used in MAE and its variants) may be suboptimal in the HOI setting. To illustrate this point, we visualize the effects of global random masking on hands and objects in Figure~\ref{fig:why_mask_stragegy}.
Intuitively, with the support of multi-view training images,
for objects with rigid structures, 
the overall structure can be inferred from a minimal amount of visible texture information (e.g., inferring the overall structure of the bottle from the visible part of the cap in the image).
However, for non-rigid hands, the situation is different.
The skin texture of human hands is similar, and each finger has a similar structure and appearance.
At the same time, hands exhibit higher complexity in kinematic performance and geometric structure.
It is challenging to infer the overall structure of the hand based solely on a small amount of visible texture information.
This leads to the problem that in traditional MAE methods,
random masking combined with a fixed high masking rate severely disrupts the structural information of the hand,
making it difficult for the model to infer the overall structure of the hand from a small amount of visible texture.

To address this issue, we propose a novel Hybrid Structure-aware Masking Strategy (Figure~\ref{fig:why_mask_stragegy}.a), which is composed of Region-specific Masking Rate Allocation (RMRA) and Skeleton-guided Sampling.

Specifically, given $\textbf{X} \in \mathbb{R}^{N \times D}$,
by guidance from the segmentation masks of hand and object and hand's keypoint localization,
RMRA is used to calculate the number of tokens to be retained for background, hand, and object. 
Specifically, 
we assign moderate, high, and low masking rates to the background, rigid objects, and non-rigid hands, 
respectively, 
based on the difficulty of learning and the importance of the target.
This dynamic masking rate allocation strategy enables the model to better adapt to the characteristics of different targets, 
thereby enhancing the feature learning ability of hands and objects in HOI scenarios.
\begin{figure*}[htbp]
    \centering
    \includegraphics[width=0.9\textwidth]{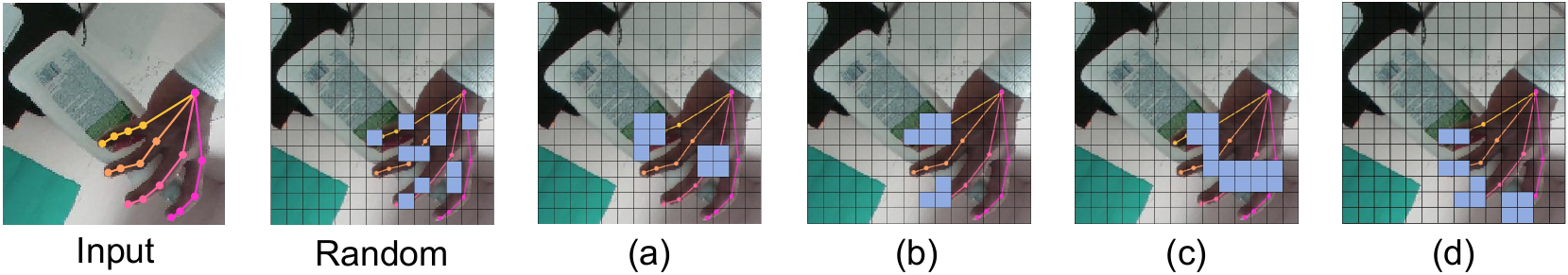}
    \caption{Visualization of the Skeleton-guided Sampling.
    }
    \label{fig:hand_mask_stragegy}
\end{figure*}

Furthermore, to adapt to the occlusion patterns in HOI scenarios,
we introduce a Skeleton-guided Sampling Strategy to replace the traditional random masking sampling for hands, which is a more challenging target.
This strategy guides the masking sampling through the hand skeleton structure, 
allowing the network to deduce the overall structure of the hand while simulating the occlusion patterns in real HOI scenarios.
Specifically, 
we design several different hand masking sampling patterns to simulate the occlusion patterns of hands in real HOI scenarios (Figure~\ref{fig:hand_mask_stragegy}), 
including but not limited to prioritizing the occlusion of a single complete finger, 
prioritizing the occlusion of all fingertips, and randomly occluding hand regions.
This structured occlusion simulation masking sampling strategy encourages our encoder to better understand the overall structure and details of the hand, 
adapting to various occlusion situations in hand-object interaction scenarios.
In particular, 
we treat the hand skeleton structure as a tree structure and perform masking sampling based on 
the hierarchical structure of the hand skeleton using both breadth-first search (BFS) and depth-first search (DFS) methods to achieve different occlusion patterns.
For rigid objects, non-rigid hands, and background, we use different masking rates based on their structural characteristics and complexity.
Finally,
 we combine the retained tokens of the three types of targets (hand, object, and background) to form a new set of unmasked tokens $\textbf{X}' \in \mathbb{R}^{N' \times D}$.

\subsection{Geometric-aware Multi-modal Learning}
\label{sec:parametric_multi_modal_learning}

As mentioned earlier,
the inherent perceptual limitations of monocular RGB inputs make it difficult for current 2D map-based HOI estimation methods to 
handle various occlusion situations commonly found in HOI scenarios.
This problem is particularly pronounced when dealing with targets like hands, which have complex shapes and variable structures.
While the previously mentioned masking strategy can alleviate occlusion issues to some extent,
it still struggles to capture occluded and self-occluded geometric information due to the inherent ambiguity of monocular input.
At the same time, due to the limitations of 2D plane-based geometric representation methods,
establishing a global geometric understanding capability of the encoder for complex HOI scenarios remains a challenge.

To address this issue, 
we propose a Geometric-aware Multi-modal Learning approach (Figure ~\ref{fig:pipeline}.c),
which drives the encoder to conduct a HOI-specific pre-training rather than a generic vision representation learning by introducing masked SDF learning methods.

In addition to the traditional 2D map supervision (segmentation masks, keypoint heatmaps) during the MAE-based training process \cite{bachmann2022multimae},
we enhance the encoder's understanding of HOI scenarios by introducing 3D Signed Distance Fields (SDF)~\cite{park2019deepsdf} as a additional modality into the learning process
to conduct Masked Signed Distance Field Learning.
SDF as a modality can enhance the encoder's understanding of HOI scenarios, in particular, going beyond the limitations of monocular input and 2D plane representation.
The introduction of this modality is primarily based on the following analysis:
(1)In HOI scenarios, there are frequent mutual occlusions between hands and objects.
Traditional monocular 2D mapping struggles to capture the occluded and self-occluded geometric information, 
and establish a global geometric representation,
while SDF can theoretically "fill in" these occluded and other perspective-visible parts, 
providing a global and more comprehensive 3D understanding.
(2) SDF, as a continuous geometric representation, provides rich information about the shape and structure of targets, 
which is crucial for understanding the geometric relationships in hand-object interactions with complex set variations.
As for hands, 
which exhibit complex variations,
SDF can provide more precise and detailed geometric information.

Specifically, as shown in Figure~\ref{fig:pipeline}.b,
after obtaining the unmasked tokens $\textbf{X}'$ from the Hybrid Structure-aware Masking Strategy,
we feed them into a transformer-based encoder to obtain the encoded tokens  and then reconstruct the masked tokens $\textbf{Y}_{h}\in \mathbb{R}^{N \times D}, \textbf{Y}_{o}\in \mathbb{R}^{N \times D}$ by two parallel decoders.
At the same time, we obtain the multi-scale feature maps $\textbf{f}_{h}, \textbf{f}_{o}$ from the decoded tokens by DPT heads~\cite{ranftl2021vision}.
Then, we obtain the SDF values $d_{h}$ and $d_{o}$ of the hand and object through sampling feature vector from the multi-scale features by projecting the random-sampled 3D query point $\textbf{p} \in \mathbb{R}^{3} $ to the 2D feature space. These process can be expressed by:

$$ d_{h} = \text{MLP}(f_{\text{FPE}}(\textbf{p})\oplus\textbf{p}\oplus\textbf{f}_{h}(\pi(\textbf{p,K})))$$
$$ d_{o} = \text{MLP}(f_{\text{FPE}}(\textbf{p})\oplus\textbf{p}\oplus\textbf{f}_{o}(\pi(\textbf{p,K})))$$

Where $\text{MLP}$ is a multi-layer perceptron, $\pi$ is the projection function, and $\textbf{K}$ is the camera intrinsic matrix, $f_{\text{FPE}}$ denotes Fourier Positional Encoding\cite{mildenhall2020nerf}.
During the pretraining process,
we optimize the model by minimizing the  L1 reconstruction loss between the predicted SDF values and the ground truth SDF values of the hand and object respectively.
The overall loss function of the masked SDF learning be expressed as $\mathcal{L} _{\text{3D}}$.
By following the previous HOI estimation methods\cite{qi2024hoisdf}, 
we compute the segmentation and hand joints heatmap loss to conduct the supervision on 2D level, 
where the loss can be expressed as $\mathcal{L}_{\text{2D}}$.

To further enhance the HOI-feature learning of our encoder,
we propose to directly use the output tokens of the network to regress the pose and MANO parameters of the hand and object, rather than relying on indirect representation methods.

There's some main reasons for this design choice:
(1) While geometric-level supervision (SDF) can enhance the encoder's understanding of scene geometry. 
However, under this condition,
the encoder tends to extract features useful for recovering geometric shapes rather than those useful for understanding pose and shape variations in hand-object interaction scenarios.
This can lead the network to generate shapes that are visually similar but physically or semantically implausible,
making it difficult to establish a direct connection between task target and encoded features.
(2) By leveraging differentiable MANO parameter model, this direct parameter regression mechanism can drive the encoder to learn the inherent structural and kinematic constraints of the human hand in MANO,
enabling the encoder to understand the intrinsic structure and kinematic constraints of the hand, rather than just learning geometric shapes.
Therefore, through this direct parameter guidance, the encoder is driven to specifically learn features crucial for determining pose and shape, rather than merely general appearance features.
This helps the encoder form a more specialized and precise understanding of the pose and shape of targets.

To achieve this particular goal, 
our architecture employs regression heads specifically chosen for the characteristics of the parameters being predicted from the decoded tokens output from their respective decoders.

Specifically, for the task of object 6D pose estimation, we utilize a simple MLP-based head for pose regression $f_{\text{pose}}$. 
For the more structurally intricate task of MANO parameter estimation, 
we employ a Transformer-based\cite{vaswani2017attention} regression head $f_{\text{MANO}}$ to effectively capture the complex articulations and dependencies of the hand.

To summarize, this process can be modified as follows:
$$(\mathbf{R}_o \in \mathbb{R}^6, \mathbf{t}_o \in \mathbb{R}^3) = f_{\text{pose}}(\mathbf{Y}_o)$$
$$(\{\mathbf{R}_h \in \mathbb{R}^6\}_{i\in{(0,16)}}, \mathbf{\beta}_h) = f_{\text{MANO}}(\mathbf{Y}_h)$$
Where $\mathbf{R}_o$ and $\mathbf{t}_o$ are the rotation and translation of the object, $\mathbf{R}_h$ and $\mathbf{\beta}_h$ are the rotation and shape parameters of the hand, respectively.
By following \cite{Wang_2021_CVPR,zhou2019continuity}, the rotations are represented as 6D vectors.
We use the L1 loss to supervise the object's pose parameters. For hand MANO parameters,
MSE loss is used to supervise parameters, the hand joints and verts loss is computed by the differentiable MANO layer\cite{romero2017embodied} in addtion.
The overall loss function of the hand and object parameters can be expressed as $\mathcal{L}_{h},\mathcal{L}_{o}$.

%% file: section/4_exp.tex
\section{Experiments}
\label{sec:experiments}
\subsection{Datasets and Metrics}

By following HOISDF, we evaluate our method on the HO3Dv2~\cite{hampali2020honnotate} and DexYCB split dataset~\cite{chao2021dexycb}~\cite{calli2015benchmarking},
and use same metrics to evaluate the performance of our method.
\subsubsection{Datasets}
For the DexYCB dataset, we use the official S0 split defined by DexYCB, which only contains the right hand images.
The pretraining and fine-tuning are conducted on the same split of DexYCB dataset.
For the HO3D dataset, we use the standard train-test splitting protocol and submit the test results to the official benchmark website.
By following HOISDF,
we used the same dataset split, containing both real and synthetic data, for both pre-training and fine-tuning, respectively.

\subsubsection{Metrics}

To evaluate our model's performance, we use the following metrics by following the previous works~\cite{qi2024hoisdf,zimmermann2017learning,xu2023h2onet,chen2022alignsdf,chen2023gsdf,wang2023interacting,hasson2019learning}:
\begin{itemize}
    \item \textbf{Hand Pose:} Mean Joint Error (MJE), Procrustes Aligned Mean Joint Error (PAMJE), Scale-Translation aligned Mean Joint Error (STMJE).
    \item \textbf{Hand Mesh Reconstruction}: Mean Mesh Error (MME), Area Under the Curve of the percentage of correct vertices (VAUC), F-scores at 5mm and 15mm (F@5mm and F@15mm). And corresponding Procrustes Aligned version.
    \item \textbf{Object Pose:} Object Center Error (OCE), Mean Corner Error (MCE), Average Closest Point Distance (ADD-S), Object Mesh Error(OME).
\end{itemize}

\begin{figure}[!htbp]
    \centering
    \includegraphics[width=1.0\columnwidth]{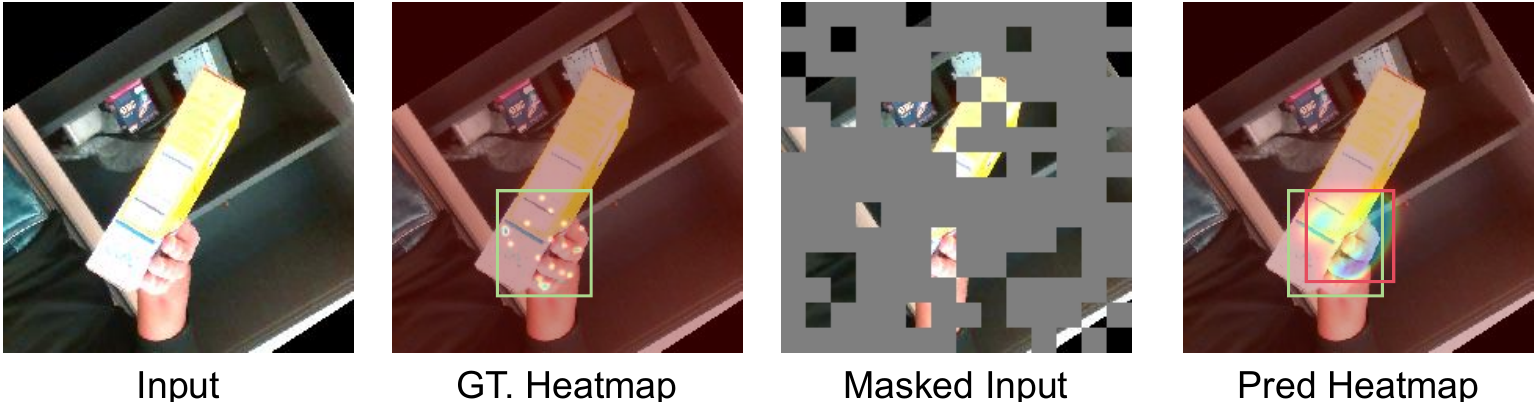}
    \caption{Predicted hand joints heatmap under global random masking.}
    \label{fig:hand_under_global_random}
\end{figure}

\begin{table*}[!htbp]
    \centering
    \begin{minipage}[t]{0.48\textwidth}
        \centering
        \input{table/comparsion_dexycb.tex}
    \end{minipage}
    \begin{minipage}[t]{0.48\textwidth}
        \centering
        \input{table/comparsion_ho3d.tex}
    \end{minipage}%
    \hfill
    \vspace{1em} 
    \begin{minipage}[t]{\textwidth}
        \centering

\input{table/comparsion_dexycb_hand.tex}
    \end{minipage}%
\end{table*}
\subsection{Implementation Details}
We use the official MAE-base pre-trained model\cite{bachmann2022multimae} with DPT head\cite{ranftl2021vision} as the U-Net-like~\cite{ronneberger2015u} backbone of our method. 
The input resolution is set to $224\times224$ for both the pretraining and fine-tuning stages, and the patch size is set to 16.
The overall loss is a weighted sum of all individual losses, which is defined as follows:
\begin{equation}
    \mathcal{L} = \lambda_{1}\mathcal{L}_{\text{2D}} + \lambda_{2}\mathcal{L}_{\text{3D}} + \lambda_{3}\mathcal{L}_{\text{h}} + \lambda_{4}\mathcal{L}_{\text{o}},
\end{equation}

Where $\lambda_{i}$ is the weight for each loss term. For the pretraining stage, 
the optimizer is AdamW~\cite{loshchilov2019decoupledweightdecayregularization} with a learning rate of 5e-5 and a weight decay of 0.05 with a batch-size of 22.
About 1k steps are used for linear warmup, and the learning rate is decayed with a cosine schedule at 72\% of the training process.
The pretraining stage conducts for 50k steps. After the pretraining, we transfer our pretrained encoder to HOISDF framework to replace the original backbone.
Then we conduct a end-to-end training process under the same setting of HOISDF to fine-tune the model. 
All experiments are implemented with PyTorch~\cite{paszke2019pytorch} and NVIDIA RTX3090 GPU.

\subsection{Comparison Experiments}

In this section, we compare our method with the state-of-the-art methods on the DexYCB split dataset and HO3D dataset.
\subsubsection{Comparison on DexYCB}Our proposed MaskHOI outperforms the state-of-the-art methods on the DexYCB split dataset, as shown in Table~\ref{tab:comparsion_dexycb}.
Compared to the baseline method HOISDF, our method achieves improvements in all metrics.
In addition, we present a more detailed comparison of hand mesh reconstruction performance in Table~\ref{tab:comparsion_dexycb_hand}.
It can be seen that our proposed MaskHOI outperforms the baseline method HOISDF in all metrics of hand, indicating the effectiveness of our method in capturing more detailed hand structures.

\begin{table*}[!htbp]
    \centering
    \begin{minipage}[t]{\textwidth}
        \centering
        \input{table/exp_ablation.tex}
    \end{minipage}%
    \hfill
    \vspace{1em} 
    \begin{minipage}[t]{0.48\textwidth}
        \centering
        \input{table/exp_ablation_mask.tex}
    \end{minipage}
    \begin{minipage}[t]{0.48\textwidth}
        \centering
        \input{table/exp_ablation_mask_ratio.tex}
    \end{minipage}%
\end{table*}
\subsubsection{Comparison on HO3D}Our proposed MaskHOI still outperforms the state-of-the-art methods on the HO3D dataset, as shown in Table~\ref{tab:comparsion_ho3d}.

Compared to the previous method\cite{lin2023harmonious}, our method only has a slight drop in PAMJE (8.9 vs 9.3), while achieving improvements in all other metrics.
Since PAMJE evaluates the joint error after alignment, this metric is more inclined to describe the hand shape estimation performance, while MJE reflects the overall performance of hand reconstruction and pose estimation.

Our method aims to propose a pre-training method and fine-tune it based on HOISDF.
By comparing PAMJE (9.6 vs 9.3), our proposed method can still effectively improve the hand reconstruction and pose estimation performance.
At the same time, the performance improvement brought by our method for object pose estimation also indicates the effectiveness of our method in hand-object interaction scenarios.

\begin{figure}[!htbp]
    \centering
    \includegraphics[width=0.8\columnwidth]{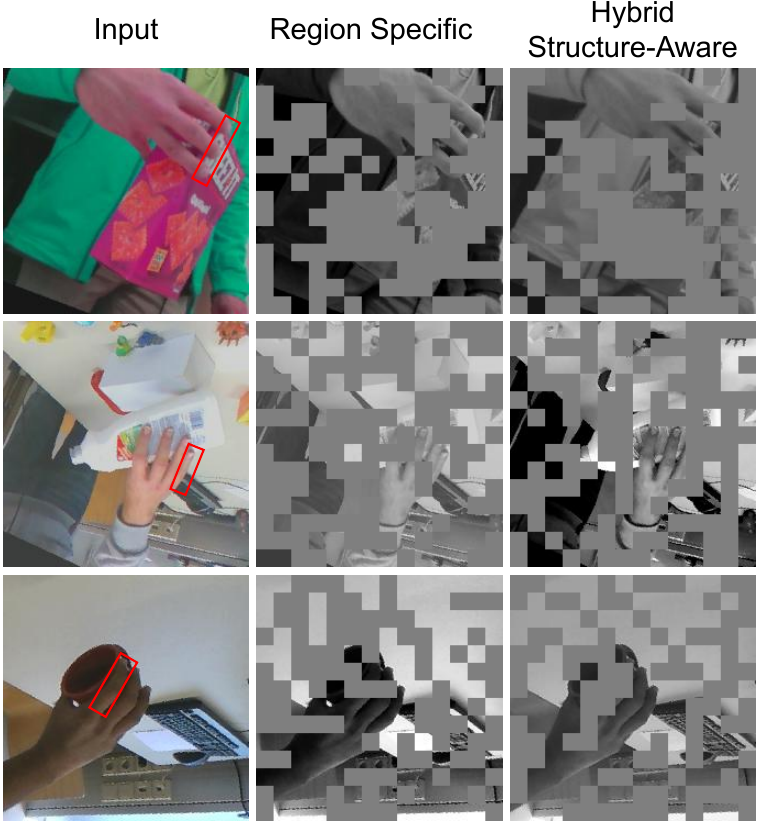}

    \caption{Images under different masking stragegy.
    Our mask strategy can adapt to the structure of the hand for targeted masking.
}
    \label{fig:mask_stragegy}
\end{figure}
\begin{figure}[!htbp]
    \centering
    \includegraphics[width=1.0\columnwidth]{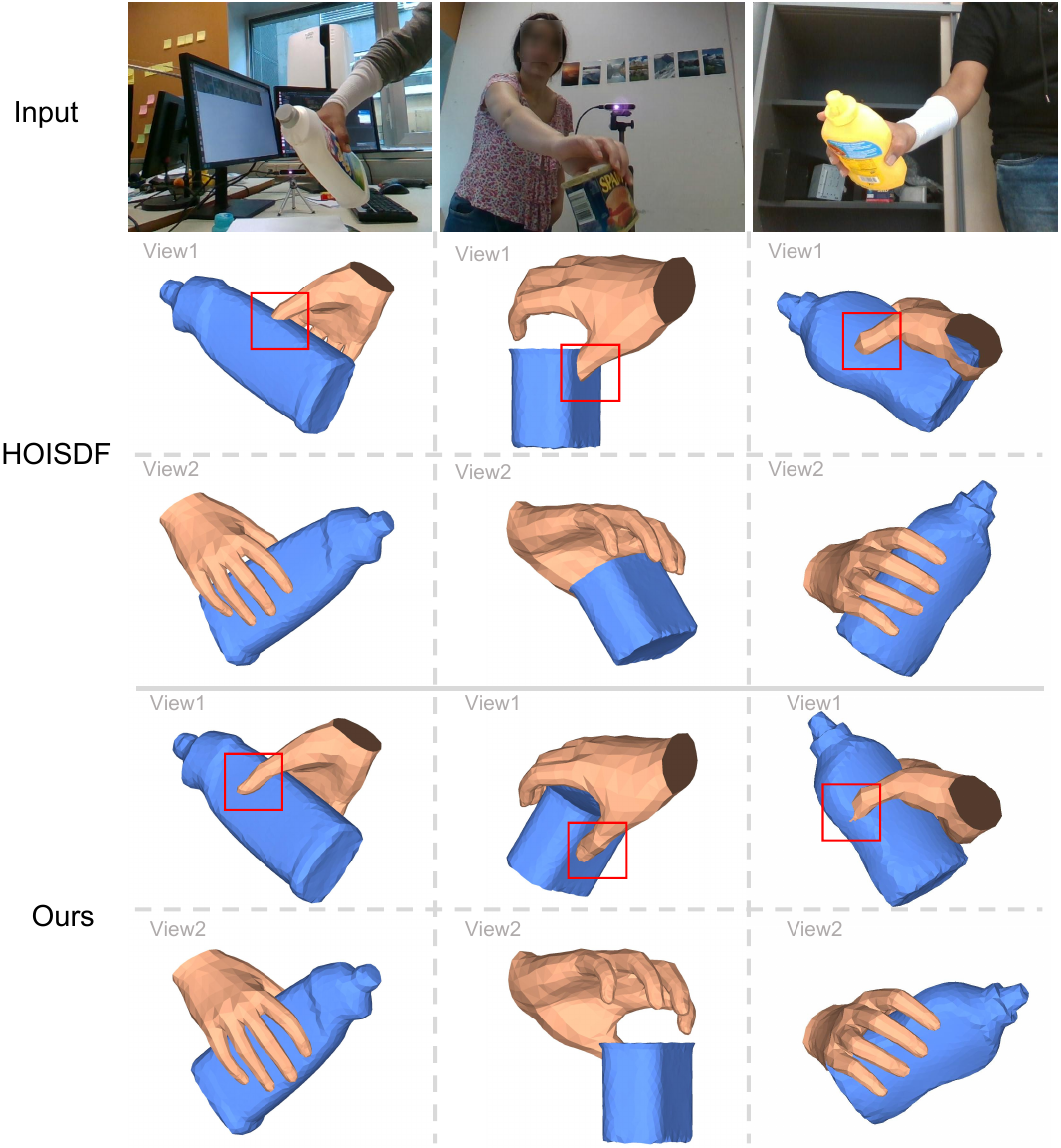}
    \caption{Qualitative comparisons on the HO3Dv2 with HOISDF.Ours method can produce more accurate estimation results under interaction scenarios.
    }
    \label{fig:q_comparsion_on_ho3d}
\end{figure}
\begin{figure*}[htbp]
    \centering
    \includegraphics[width=0.9\textwidth]{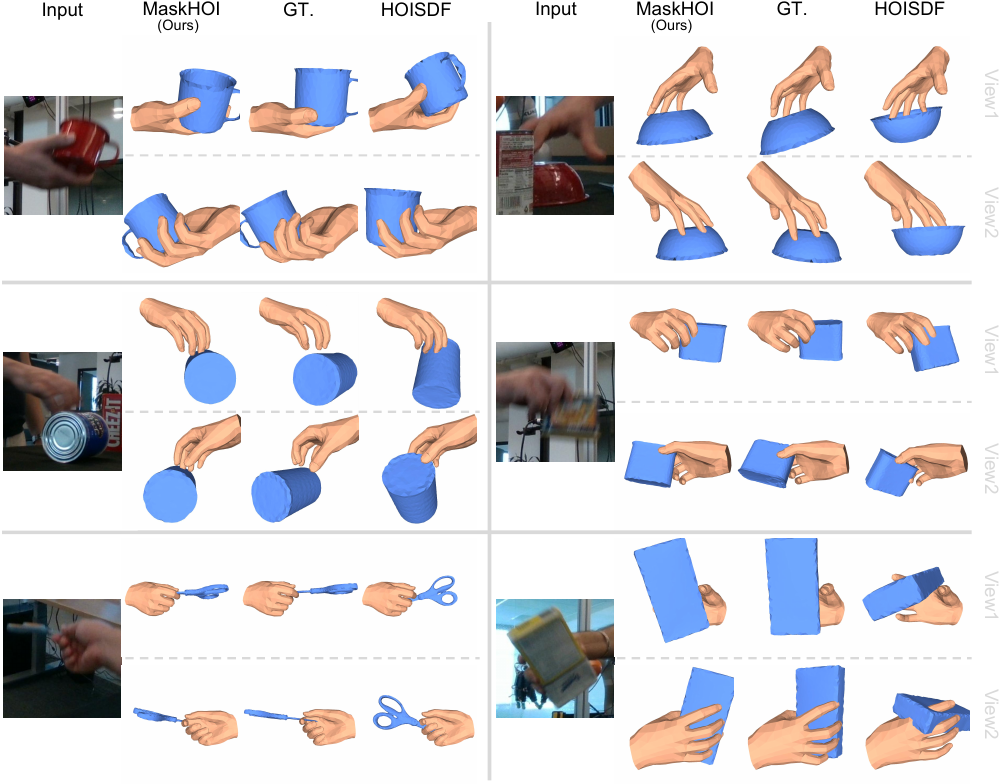}
    \caption{Qualitative comparisons on the DexYCB with HOISDF.Ours method can produce more accurate estimation results under interaction scenarios.
    }
    \label{fig:q_comparsion_on_DexYCB}
\end{figure*}
\subsection{Ablation Study}

\subsubsection{Effectiveness of key components}
To verify the effectiveness of each component in our method, we conduct an ablation study on the HO3D dataset.
The results are shown in Table~\ref{tab:exp_ablation_mask_hoi}.
To ensure a fair comparison, we present the results of fine-tuning with the official MAE pre-trained weights\cite{he2022masked} in the second row of the table.
Here, GML, DPR, and HSMS represent the use of 
Geometric-aware Multi-modal Learning , 
Direct Parameters Regression, 
and Hybrid Structure-aware Mask Strategy, respectively.

\subsubsection{Effectiveness of Mask Strategy}

We show the predicted results under global random masking in Figure~\ref{fig:hand_under_global_random}.
Intuitively, it is difficult to infer the global information of the hand from the masked image.
By comparing the information within the red and green boxes, we can see that the severe information loss leads to an overall offset in the predicted joint positions of the hand.
This misleads the network optimization, potentially causing a performance drop.

Therefore, to verify the effectiveness of our designed Hybrid Structure-aware Mask Strategy, we conduct an ablation study on the HO3D dataset.
The results are shown in Table~\ref{tab:exp_ablation_mask}.

We compare three different mask strategies, including Global Random mask, Region-specific Masking Rate Allocation, and the Hybrid Structure-aware mask.
The results show that our designed Hybrid Structure-aware Mask Strategy can effectively improve the model's performance in hand-object interaction scenarios.

Meanwhile, we adjust the applied proportion of Skeleton-guided in the Hybrid Structure-aware Mask Strategy to avoid potential information leakage of hand structure caused by Skeleton guidance, and conduct an ablation study on it, as shown in Table~\ref{tab:exp_ablation_mask_ratio}.

\subsection{Qualitative Comparisons}
We conduct qualitative comparisons on the DexYCB split dataset and HO3D dataset, as shown in Figure ~\ref{fig:q_comparsion_on_ho3d} and Figure~\ref{fig:q_comparsion_on_DexYCB}.

The results show that our method achieves better performance in HOI scene estimation under difficult settings which contain occlusion, self-occlusion, strong dynamic blur, and slender-shaped objects.

By comparing with HOISDF, our method achieves better performance in diffcult HOI scene estimation under occlusion, self-occlusion, and strong dynamic blur scenarios.

This indicates that our method can better capture detailed information in hand-object interaction scenarios.

So that our proposed MaskHOI can generate more reasonable scene estimation results.

%% file: table/comparsion_dexycb.tex
    \centering
    \caption{Comparsion experiments on DexYCB split dataset.}
    \label{tab:comparsion_dexycb}
    \huge
    \resizebox{\columnwidth}{!}{  
        \begin{tabular}{l|cc|cccc}
            \toprule

            \multirow{2}{*}{\textbf{METHOD}} & \multicolumn{2}{c|}{\textbf{HAND}} & \multicolumn{3}{c}{\textbf{OBJ}} \\
            \cmidrule(lr){2-3} \cmidrule(lr){4-6}
            & \textbf{MJE}($\downarrow$) & \textbf{PAMJE}($\downarrow$) & \textbf{OCE}($\downarrow$) & \textbf{MCE}($\downarrow$) & \textbf{ADD-S}($\downarrow$)\\
            \midrule
            Hasson \textit{et al.}\cite{hasson2019learning}  & 17.6 & - & - & - & -  \\
            Hasson \textit{et al.}\cite{hasson2020leveraging}  & 18.8 & - & - & 52.5 & -  \\
            Tze \textit{et al.}\cite{tse2022collaborative}  & 15.3 & - & - & - & -  \\
            Li \textit{et al.}\cite{li2021artiboost}  & 12.8 & - & - & - & -  \\
            Chen \textit{et al.}\cite{chen2022alignsdf}  & 19.0 & - & 27.0 & - & -  \\
            Chen \textit{et al.}\cite{chen2023gsdf}  & 14.4 & - & 19.1 & - & -  \\
            Lu \textit{et al.}~\cite{lu2024spmhand}  & 13.6 & 5.50 & - & - & - \\
            Wang \textit{et al.}\cite{wang2023interacting}  & 12.7 & 6.86 & 27.3 & 32.6 & 15.9  \\
            Lin \textit{et al.}\cite{lin2023harmonious}  & 11.9 & 5.81 & 39.8 & 45.7 & 31.9  \\
            HOISDF\cite{qi2024hoisdf} & 10.1 & 5.31 & 18.4 &27.4 & 13.3  \\
            \midrule
            MaskHOI (ours) &\textbf{10.0} 	&\textbf{5.07} 	&\textbf{16.7} 	&\textbf{25.3} 	&\textbf{12.3}             \\
            \bottomrule
            \end{tabular}}

%% file: table/comparsion_ho3d.tex
    \centering
    \caption{Comparsion experiments on HO3D dataset. "*" denoted models that were co-trained with synthetic data.}
    \label{tab:comparsion_ho3d}
    \huge 
    \resizebox{\columnwidth}{!}{  
    \begin{tabular}{l | ccc | cc}
    \toprule

    \multirow{2}{*}{\textbf{METHOD}} & \multicolumn{3}{c|}{\textbf{HAND}} & \multicolumn{2}{c}{\textbf{OBJ}} \\
    \cmidrule(lr){2-4} \cmidrule(lr){5-6}
    & \textbf{MJE}($\downarrow$) & \textbf{STMJE}($\downarrow$) & \textbf{PAMJE}($\downarrow$) & \textbf{OME}($\downarrow$) & \textbf{ADD-S}($\downarrow$) \\
    \midrule
    Hasson \textit{et al.}~\cite{hasson2019learning}   & ---   & ---   & 31.8 & 11.0 & ---   \\
    Hasson \textit{et al.}~\cite{hasson2020leveraging}   & ---   & 36.9  & 11.4 & 67.0 & 22.0  \\
    Hasson \textit{et al.}~\cite{hasson2021towards}   & ---   & 26.8  & 12.0 & 80.0 & 40.0  \\
    Liu \textit{et al.}~\cite{liu2021semi}      & ---   & 31.7  & 10.1 & ---  & ---   \\
    Hampali \textit{et al.}~\cite{hampali2022keypoint} & 25.5  & 25.7  & 10.8 & 68.0 & 21.4  \\
    Lin \textit{et al.}~\cite{lin2023harmonious}      & 28.9  & 28.4  & \textbf{8.9}  & 64.3 & 32.4  \\
    Lu \textit{et al.}~\cite{lu2024spmhand}        & 24.1  & -  & 9.0  & - & -  \\
    HOISDF~\cite{qi2024hoisdf}                   & 23.6  & 22.8  & 9.6  & 48.5 & 17.8  \\
    MaskHOI (ours)          & \textbf{19.7}  & \textbf{19.0}  & 9.3  & \textbf{45.6} & \textbf{16.1}  \\
    \midrule 
    Li \textit{et al.}*~\cite{li2021artiboost}       & 26.3  & 25.3  & 11.4 & ---  & ---   \\
    Wang \textit{et al.}*~\cite{wang2023interacting}     & 22.2  & 23.8  & 10.1 & 45.5 & 20.8  \\
    HOISDF*~\cite{qi2024hoisdf}                  & 19.0  & 18.3  & 9.2  & 35.5 & 14.4  \\
    MaskHOI* (ours)          & \textbf{18.0} & \textbf{17.5} & \textbf{8.9} & \textbf{33.6} & \textbf{13.1} \\
    \bottomrule
    \end{tabular}}

%% file: table/comparsion_dexycb_hand.tex
    \centering
    \caption{Comparison with hand mesh metrics on the DexYCB split testset. MME and PAMME are in millimeters.}
    \label{tab:comparsion_dexycb_hand}
    \begin{tabular}{ l c c c c c c c c}
        \toprule
        Metrics & MME $\downarrow$ & VAUC $\uparrow$ & F@5 $\uparrow$ & F@15 $\uparrow$ & PAMME $\downarrow$ & PAVAUC $\uparrow$ & PAF@5 $\uparrow$ & PAF@15 $\uparrow$ \\
        \hline
        HandOccNet\cite{park2022handoccnet} &-  &76.6 &51.5  &92.4  &-  &89.0  &78.0  &99.0  \\
        H2ONet\cite{xu2023h2onet} &-  &76.2 &51.3  &92.1  &-  &89.1  &80.1  &99.0  \\
        HOISDF\cite{qi2024hoisdf} & 10.1 & 80.1 & 58.7 & 94.6 & 5.1 & 89.7 & 80.4 & 99.1 \\
        MaskHOI (ours) & \textbf{9.7} & \textbf{80.8} & \textbf{59.9} & \textbf{95.1} & \textbf{4.9} & \textbf{90.3} & \textbf{82.2} & \textbf{99.3} \\
        \bottomrule
    \end{tabular}

%% file: table/exp_ablation.tex
    \centering

    \caption{Ablation experiments of key component of MaskHOI on HO3D dataset.}
    \label{tab:exp_ablation_mask_hoi}
    \begin{tabular}{c c c c c c c c c c}
    \toprule
    \multirow{2}{*}{\textbf{METHOD}} & \multicolumn{3}{c}{\textbf{Setting}} & \multicolumn{3}{c}{\textbf{HAND}} & \multicolumn{2}{c}{\textbf{OBJ}} & \multirow{2}{*}{\textbf{MEAN}} \\
    \cmidrule(r){2-4} \cmidrule(r){5-7} \cmidrule(r){8-9}
    & \textbf{GML} & \textbf{DPR} & \textbf{HSMS} & \textbf{MJE($\downarrow$)} & \textbf{STMJE($\downarrow$)} & \textbf{PAMJE($\downarrow$)} & \textbf{OME($\downarrow$)} & \textbf{ADD-S($\downarrow$)} & \\
    \midrule
    HOISDF & $\times$ & $\times$ & $\times$ & 23.6 & 22.8 & 9.6 & 48.5 & 17.8 &24.5     \\
    \midrule
    \multirow{4}{*}{Ours} &\multicolumn{3}{c}{[Vanilla MAE Init]} &22.2 & 21.6 & 9.9 & 50.1 & 17.7 &24.3 \\
    & $\checkmark$ & & & 21.9 & 21.1 & 10.1 & 47.4 & \textbf{14.9}  &23.1 \\
    & $\checkmark$ & $\checkmark$ & & 21.8 & 21.0 & 9.9 & \textbf{44.0} & 15.9 &22.5\\
    & $\checkmark$ & $\checkmark$ & $\checkmark$ & \textbf{19.7} & \textbf{19.0} & \textbf{9.3} & 45.6 & 16.2 &\textbf{22.0} \\
    \bottomrule
    \end{tabular}

%% file: table/exp_ablation_mask.tex
\centering

\caption{Ablation study of components of Hybrid Structure-aware Masking Strategy on HO3D dataset.}
\label{tab:exp_ablation_mask}
\resizebox{\columnwidth}{!}{
\begin{tabular}{cc  ccc  cc}
    \toprule
    \multicolumn{2}{c}{\textbf{Setting}} & \multicolumn{3}{c}{\textbf{HAND}} & \multicolumn{2}{c}{\textbf{OBJ}}  \\
    \cmidrule(lr){1-2} \cmidrule(lr){3-5} \cmidrule(lr){6-7}
    \textbf{RMRA} &\textbf{SG} & \textbf{MJE}($\downarrow$) & \textbf{STMJE}($\downarrow$) & \textbf{PAMJE}($\downarrow$) & \textbf{OME}($\downarrow$) & \textbf{ADD-S}($\downarrow$) \\
    \midrule
    $\times$           &$\times$    &21.8 	&21.0 	&9.9 	&\textbf{44.0} 	&15.9     \\
    $\checkmark$       &$\times$    &20.6 	&19.9 	&9.5 	&44.9 	&\textbf{15.8}     \\
    $\checkmark$       &$\checkmark$    &\textbf{19.7} 	&\textbf{19.0} 	&\textbf{9.3} 	&45.6 	&16.2    \\
    \bottomrule
\end{tabular}
}

%% file: table/exp_ablation_mask_ratio.tex
\centering

\caption{
    Ablation study on the applied proportion of the Skeleton guidance within the Hybrid Structure-aware Masking Strategy.
    The first row indicates a gradual increase in the proportion during the training process.
}
\label{tab:exp_ablation_mask_ratio}
\resizebox{\columnwidth}{!}{
\begin{tabular}{c  ccc cc}
    \toprule
    \multirow{2}{*}{\textbf{Proportion}} & \multicolumn{3}{c}{\textbf{HAND}} & \multicolumn{2}{c}{\textbf{OBJ}}  \\
     \cmidrule(lr){2-4} \cmidrule(lr){5-6}
      & \textbf{MJE}($\downarrow$) & \textbf{STMJE}($\downarrow$) & \textbf{PAMJE}($\downarrow$) & \textbf{OME}($\downarrow$) & \textbf{ADD-S}($\downarrow$) \\
    \midrule
    0.0 $\rightarrow$ 1.0           &20.9 	&20.9 	&9.4 	&46.1 	&16.7     \\
    1.0                             &20.6 	&19.9 	&9.9 	&45.7 	&\textbf{15.7}     \\
    0.0                             &20.6 	&19.9 	&9.5 	&\textbf{44.9} 	&15.8     \\
    0.5                             &\textbf{19.7} 	&\textbf{19.0} 	&\textbf{9.3} 	&45.6 	&16.2     \\
    \bottomrule
\end{tabular}
}

%% file: section/6_conclusion.tex
\section{Conclusion}
\label{sec:conclusion}

In this paper, we propose a novel Masked Autoencoder (MAE)–driven pretraining framework for enhanced HOI pose estimation. 
The core idea of our method is to leverage the masking-then-reconstruction strategy of MAE 
to encourage the feature encoder to infer missing spatial and structural information, 
thereby facilitating geometric-aware and occlusion-robust representation learning. 
Additionally, by designing a Hybrid Structure-aware Masking Strategy and a global Geometric-aware Multimodal Learning mechanism,
the model can better capture the complex interaction relationships between the hand and object, thereby improving the accuracy and robustness of HOI estimation.